\documentclass[conference]{IEEEtran}
\IEEEoverridecommandlockouts
\usepackage{cite}
\usepackage{amsmath,amssymb,amsfonts}
\usepackage{algorithmic}
\usepackage{graphicx}
\usepackage{textcomp}
\usepackage{xcolor}
\usepackage{float}
\usepackage{url}
\usepackage{subfigure}
\def\BibTeX{{\rm B\kern-.05em{\sc i\kern-.025em b}\kern-.08em
    T\kern-.1667em\lower.7ex\hbox{E}\kern-.125emX}}
\begin{document}

\title{Federated Learning for Tabular Data using TabNet: A Vehicular Use-Case\\
}

\author{\IEEEauthorblockN{William Lindskog}
\IEEEauthorblockA{\textit{Corporate Research and Development} \\
\textit{DENSO Automotive Deutschland GmbH}\\
Munich, Germany \\
w.lindskog@eu.denso.com}
\and
\IEEEauthorblockN{Christian Prehofer}
\IEEEauthorblockA{\textit{Corporate Research and Development} \\
\textit{DENSO Automotive Deutschland GmbH}\\
Munich, Germany \\
c.prehofer@eu.denso.com}
}
\IEEEoverridecommandlockouts
\IEEEpubid{\makebox[\columnwidth]{978-1-6654-6437-6/22/\$31.00~\copyright2022 IEEE \hfill} \hspace{\columnsep}\makebox[\columnwidth]{ }}
\maketitle
\IEEEpubidadjcol
\begin{abstract}
In this paper, we show how Federated Learning (FL) can be applied to vehicular use-cases in which we seek to classify obstacles, irregularities and pavement types on roads. Our proposed framework utilizes FL and TabNet, a state-of-the-art neural network for tabular data. We are the first to demonstrate how TabNet can be integrated with FL. Moreover, we achieve a maximum test accuracy of 93.6$\%$. Finally, we reason why FL is a suitable concept for this data set. 
\end{abstract}


\section{Introduction}\label{sec:Introduction}
Federated Learning (FL) is a collaborative machine learning concept which advocates local computing and model transmission. Instead of sending raw data to central (or partially central) servers for computing, FL ensures that data are kept on the edge nodes and can reduce overhead communication. Google developed FL as a countermeasure to regulators' incentives to increase protection of consumer data \cite{konevcny2016federated}. Since then, research has mainly focused on the theoretical foundations of the concept, its optimization algorithms, and comparing it to conventional machine learning on standard data sets. 

\begin{figure}[h!]
    \centering
    \includegraphics[width = 7cm]{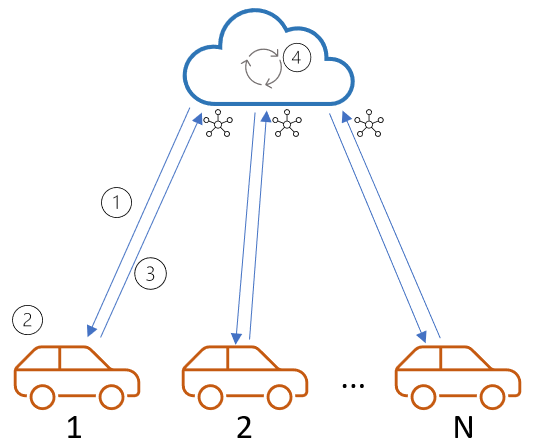}
    \caption{A FL vehicular scheme: (1) Server sends out initial model, (2) vehicles train model on local data, (3) send back model updates, and (4) server aggregates model updates. This concludes a FL round.}
    \label{fig:vehicle_FL_scheme}
\end{figure}

FL as a technique is starting to gain traction in various industries e.g. healthcare\cite{rieke2020future, pfitzner2021federated}, internet-of-things (IoT)\cite{khan2021federated, nguyen2021federated, imteaj2021survey, kholod2020open}, and lately Intelligent Connected Vehicles (ICVs)\cite{du2020federated, posner2021federated}, see Figure \ref{fig:vehicle_FL_scheme} for conventional vehicular FL scheme. FL for ICVs has seen great improvement in its system design where researchers propose peer-to-peer networks \cite{lu2020federated}, vehicle-to-everything communication\cite{zhou2021two}, and network optimization with asynchronous update schemes \cite{zhang2021af}. Research has also identified FL as a suitable concept for vehicular networks and fleet management \cite{posner2021federated, lim2020federated}. Despite the extensive work being conducted in FL research, it lacks real-world implementations. Applied FL is a step in the direction of protecting end-user data. By collaboratively training a shared model, many cooperating ICVs can encounter more events than a single vehicle, and potentially train a superior model.

\begin{figure}[h!]
    \centering
    \includegraphics[width = 9cm]{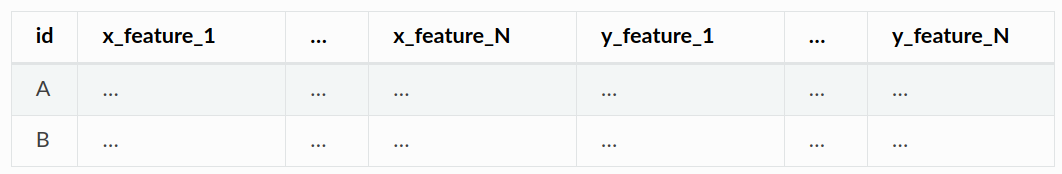}
    \caption{Example list of features in tabular form \cite{christ2018time}}
    \label{fig:feature_list}
\end{figure}

Research that seeks to apply FL to vehicles generally centers around image data. A vehicle produces a plethora of data, much of which can be seen as time series data e.g. acceleration, velocity, engine temperature. This data can be used to improve consumer comfort, and to increase safety by applying accurate fault diagnosis and predictive maintenance. In IoT, we see that research is heavily invested in time series analysis, where FL is now beginning to be integrated \cite{liu2020deep, kholod2020open}. 
Applying feature extraction on time series data has proven to be advantageous for time series classification, resulting in work with tabular data \cite{del2021active}, see Figure \ref{fig:feature_list} for example. 

In search for suitable algorithms for tabular data, we find that Google's recent TabNet algorithm\cite{arik2021tabnet} can be superior to other state-of-the-art methods e.g. Random Forests and XGBoost (which already have FL implementations). \cite{arik2021tabnet} show that TabNet outperforms the other algorithms in regression and classification tasks. However, TabNet is yet to be combined with FL, and we demonstrate how this can be done. In our work, we therefore construct a framework, in which both TabNet and FL can be used together. We apply a feature extraction pre-processing step on time series to work with tabular data. We limit our work to TabNet and FL, and compare it with other studies. Our direct contributions to research are:
 (1) Integrating TabNet with FL, (2) showing that TabNet and FL can achieve similar performance to state-of-the-art methods, and (3) a conceptual framework combining FL and TabNet that can be used for tabular data and converted time series to tabular data.

\section{Data}\label{sec:Data}
\begin{figure}[h!]
    \centering
    \includegraphics[width = 8cm]{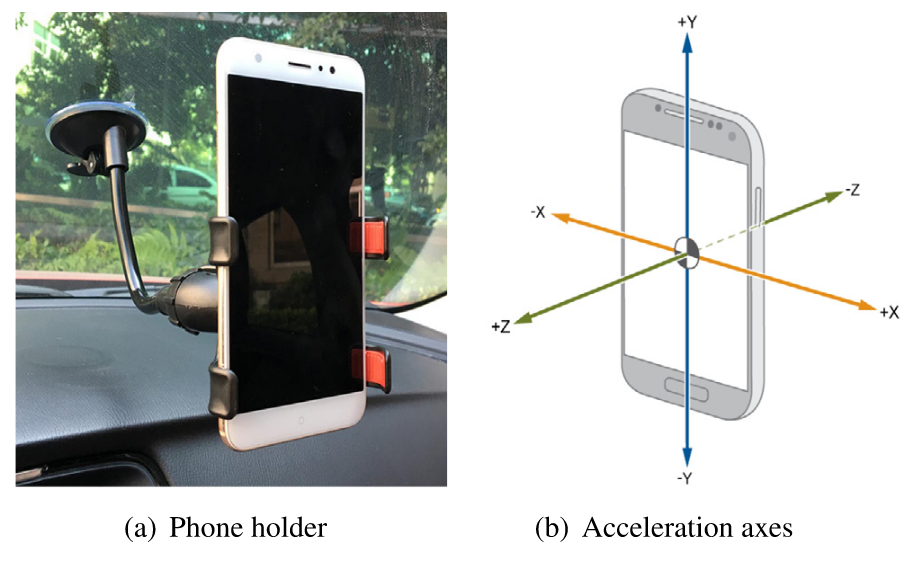}
    \caption{Example of car phone holder used to collect data and directions
considered by the accelerometer sensor \cite{souza2018asphalt}}
    \label{fig:smartphone}
\end{figure}

The data we use, have been collected by \cite{souza2018asphalt}. By installing a smartphone inside the vehicle cabin, as in Figure \ref{fig:smartphone}, they registered the trilateral acceleration for further analysis of the asphalt/road conditions. The trilateral accelerations are given by an accelerometer, integrated in the smartphone, which continuously collects time series data. The sampling rate that they use is 100Hz. 

In their paper, they divide and specify three distinct classification tasks for which they allocate independent labelled data sets, which we also consider for our work. 

\begin{table}[h!]
    \centering
    \caption{Class distribution of time series for the three data sets}
    \tabcolsep=0.09cm
    \begin{tabular}{llccccc}
    \hline
    \textbf{Data set} & \textbf{Class} & \multicolumn{3}{c}{\textbf{Series Length}} & \textbf{Examples} & \textbf{Distribution} \\
    \cline{3-5}
    & & Min & Max & Mean \\
    \hline \hline
    \textbf{Regularity}& Regular&66&2371&238&762&50.73$\%$ \\
    &Deteriorated&190&4201&534&740&49.27$\%$ \\
    \hline
    \textbf{Pavement}&Flexible&66&2371&246&816&38.65$\%$\\
    \textbf{Type}&Cobblestone&284&1543&518&527&24.97$\%$\\
    &Dirt Road&274&1045&484&768&36.38$\%$\\
    \hline
    \textbf{Obstacles}&Speed Bump&178&730&330&212&27.14$\%$\\
    &Vertical Patch&114&279&191&222&28.43$\%$\\
    &Raised Markers&111&462&256&187&23.94$\%$\\
    &Raised Crosswalk&258&736&457&160&20.49$\%$\\
    \hline
    \end{tabular}
    \label{tab:datasets}
\end{table}

\begin{figure}[h!]
    \centering
    \includegraphics[width = 8cm]{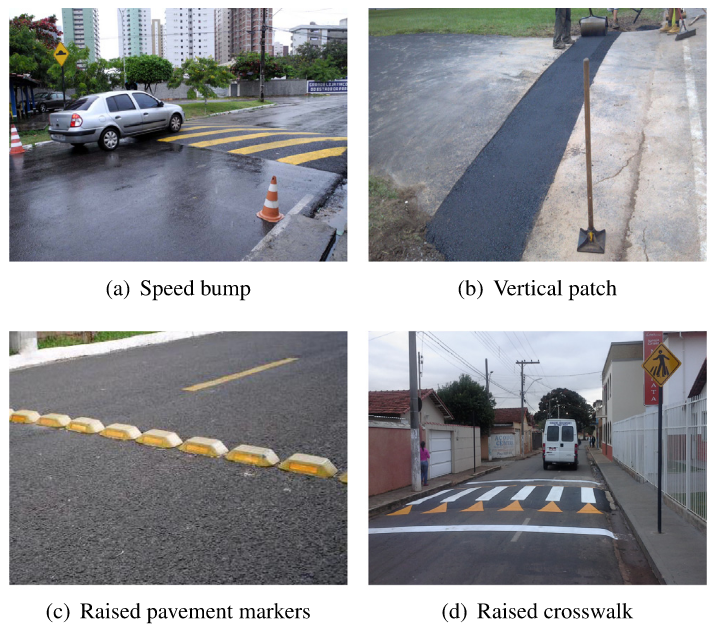}
    \caption{Photos that illustrate the four class labels of the Asphalt-Obstacles
data set \cite{souza2018asphalt}}
    \label{fig:obstacles}
\end{figure}

The first data set is the \textit{Asphalt Regularity} data set. It considers two classes: (1) Regular pavement where driver comfort is high and (2) deteriorated where they observe irregularities and roughness in the road. Second, the \textit{Pavement Type} data set. \cite{souza2018asphalt} identifies three pavement type classes: (1) Flexible pavement, (2) cobblestones, and (3) dirt roads. Flexible pavements are defined as a mixture of asphalt and bituminous material. The final data set, \textit{Asphalt Obstacles}, includes common obstacles on roads with specific characteristics, see Figure \ref{fig:obstacles}. In their study, they identify a great variability for examples of the same class (width, height, size, and material). 

\section{Method}\label{sec:Method}
Initially we manually extract features from all time series using the work from \cite{christ2018time}. Thereafter, the high level architecture of our processing scheme can be described shortly as: 
\begin{enumerate}
    \item Distribute copies of initial server model to clients (nodes).
    \item Train local model on local training set. 
    \item Add new instances to local training set from local pool of instances to reflect real-world.
    \item Validate local model on local validation set.
    \item Transmit model updates to server. 
    \item Server aggregates model updates and evaluates on test set.  
\end{enumerate}
On the server-side, we therefore keep one test set which is used to evaluate the aggregated server model after each communication round, where a communication round is illustrated in Figure \ref{fig:vehicle_FL_scheme}.
On each client, we keep three local data sets:
\begin{enumerate}
    \item (Local) Training set, grows in size over elapsed processing time. The reason why it grows is because a real-world operating vehicle would experience more situations as it goes. 
    \item (Local) Pool set from which we draw random samples from and add to local training set, if pool instances still exist for respective client. 
    \item Validation set, for tuning hyperparameters in TabNet. 
\end{enumerate}
We randomly select a small arbitrary number of instances and add to add them to the local training set. We continue this process until there are no more instances in the local pool set. The FL environment is constructed using Flower\cite{beutel2020flower}. It is an end-to-end FL framework, designed to enable a seamless transition from experimental research to real edge devices. We set up the FL environment on one machine using Flower's built-in Virtual Client Engine (VCE) that enables the virtualization of Flower Clients to maximise utilization of the available hardware. Moreover, we choose to use FedAvg \cite{konevcny2016federated} as FL optimization algorithm. 

\subsection{TabNet}\label{subsec:TabNet}
TabNet is based on the idea of representing a deep neural network as decision trees (DTs). DTs (and other similar algorithms e.g. XGBoost \cite{chen2015xgboost} and Random Forests) are known to perform well on tabular data. It uses tabular input as data for classification- or self-supervised tasks and applies sparse features selection which can enable interpretability and better learning \cite{li2020survey}. The use of sparse feature selection includes processing a subset of features in step-wise order, and later aggregating information from subset feature processing. The selected features are linearly transformed and together with additional bias they form TabNet's decision boundaries. TabNet is trained using gradient decent-based optimization and it incorporates an encoder-decoder architecture \cite{arik2021tabnet}. 

\section{Evaluation and Results}\label{sec:Evaluation_and_Results}
For our evaluation, we use an Alienware, Ubuntu 21.04, 16core CPU, Intel Core i7-10870H processor at 2.20GHz clock speed, and a 16GB NVIDIA GeForce RTX 3080 GPU. 

When configuring TabNet's hyperparameters, we follow the guidelines in their article \cite{arik2021tabnet}. Nevertheless, we recognize that they work with larger data sets than ours. Therefore, our base settings differ from what they advocate, mainly in terms of width of decision prediction layer and attentive embedding. We use a value of five for both of these values, while \cite{arik2021tabnet} recommend a value between 8-64. Our choice of lower values than recommended is motivated by initial attempts to tune TabNet's hyperparameters. For our data sets, it seems that a slight increase in the width of decision prediction layer and attentive embedding leads to overfitting the model. Everything else we leave as default setting, presented in their ablation study. We choose to evaluate the federated set-up using 2-3 clients. When adding more than three clients to the FL set-up, our machine sometimes runs out of GPU memory. 

Section \ref{sec:Evaluation_and_Results} is divided into three parts, one for each data set. For each data set, we present test accuracy and test cross-entropy loss over 100 communication rounds. The accuracy and loss is computed on the server-side after one communication round and aggregation of model updates. Furthermore, we also include confusion matrices for the data sets with more than two classes. In a two-client set-up, we wait for each client to send its updates to the server. Using three clients, we sample two out of three to send their updates. Lastly, we split each data set into four subsets: 
\begin{enumerate}
    \item (Training set) $10\%$ of total data, equally distributed to clients. Each client's training set increase with 10 instances per communication round.   
    \item (Pool set) $60\%$ of total data, equally distributed to clients. Each client's pool set decreases with 10 instances per communication round.
    \item (Validation set) $10\%$ of total data, equally distributed to clients. 
    \item (Test data set) $20\%$ of total data, kept on the server for evaluation. 
\end{enumerate}
\subsection{Asphalt Regularity}\label{subsec:asphalt_regularity(RESULTS)}
The first data set is a balanced data set with two classes (see Table \ref{tab:datasets}). The server sends out an initial models to each client. The clients sample random instances from their respective pool set and adds them to the training set for the next local training round. Eventually the clients run out of samples from the pool set but the training continues. This is also true for the other two data sets. 

\begin{figure*}[h!]
    \centering
    \subfigure[Test accuracy for asphalt regularity data set.]{\includegraphics[width = 8.7cm, height = 6.0cm]{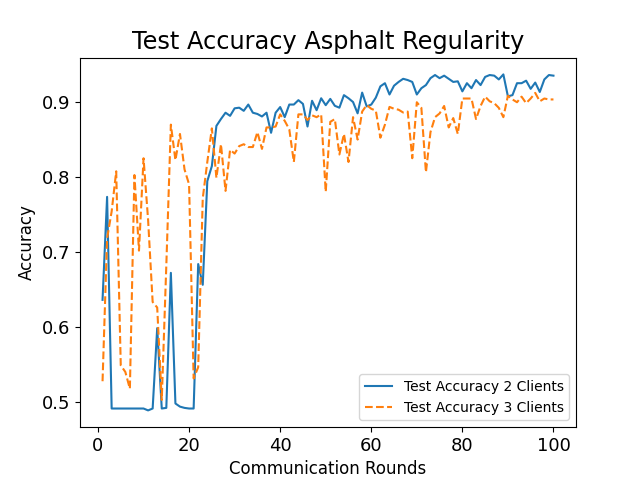}}\quad
    \subfigure[Test loss for asphalt regularity data set.]{\includegraphics[width = 8.7cm, height = 6.0cm]{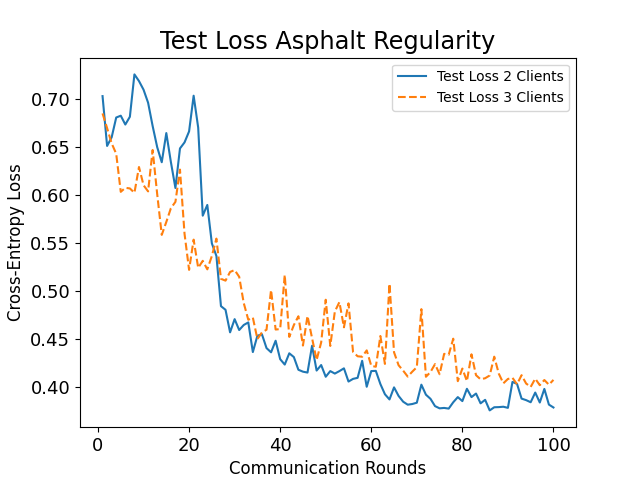}}
    \caption{Performance of asphalt regularity data set.}
    \label{fig:performance_asphalt_regularity}
\end{figure*}


We illustrate the test accuracy results and cross entropy loss in Figure \ref{fig:performance_asphalt_regularity}. The maximum test accuracy reached is 93.6$\%$ and we achieve this with two clients. Adding an additional client reduces maximum test accuracy slightly to 91.2$\%$. The test accuracy is greatly varying until approximately communication round 30. Thereafter, it seems to increase linearly until communication round 70-80 and later converges towards maximum test accuracy.  
\subsection{Pavement Type}\label{subsec:pavement_type(RESULTS)}
\begin{figure*}[h!]
    \centering
    \subfigure[Test accuracy for pavement type data set.]{\includegraphics[width = 8.7cm, height = 6.0cm]{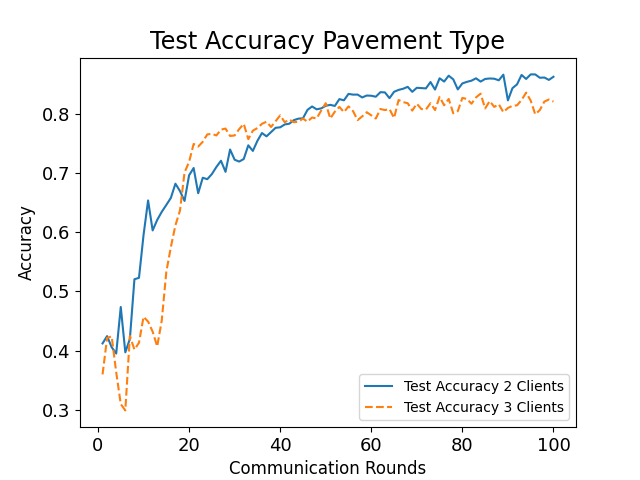}}\quad
    \subfigure[Test loss for pavement type data set.]{\includegraphics[width = 8.7cm, height = 6.0cm]{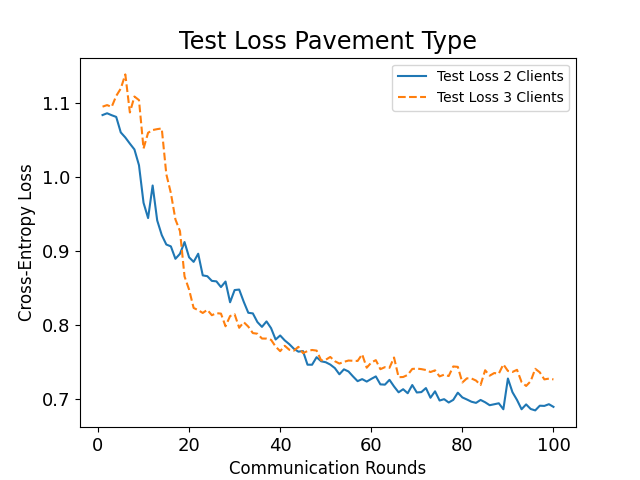}}
    \caption{Performance of pavement type data set.}
    \label{fig:performance_pavement_type}
\end{figure*}

The second data set is less balanced than the first one. There number of time series stemming from cobblestones are significantly less than the other two classes: (1) flexible and (2) dirt road. For this data set, we also include a confusion matrix to better illustrate correct and incorrect predictions. The confusion matrix shows the predictions the server in a two-client set-up since we achieve maximum test accuracy with two clients.  Maximum test accuracy of 86.7$\%$ is achieved, again with two clients. For three clients, maximum test accuracy is 83.6$\%$, and this is shown in Figure \ref{fig:performance_pavement_type}.

\begin{figure*}[h!]
    \centering
    \subfigure[Test accuracy for asphalt obstacles data set.]{\includegraphics[width = 8.7cm, height = 6.0cm]{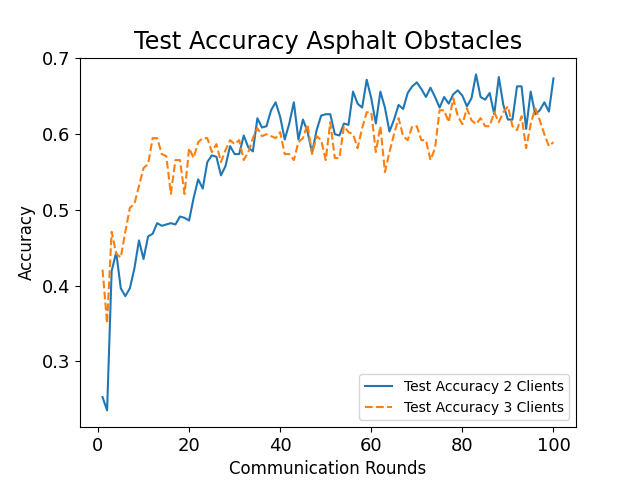}}\quad
    \subfigure[Test loss for asphalt obstacles data set.]{\includegraphics[width = 8.7cm, height = 6.0cm]{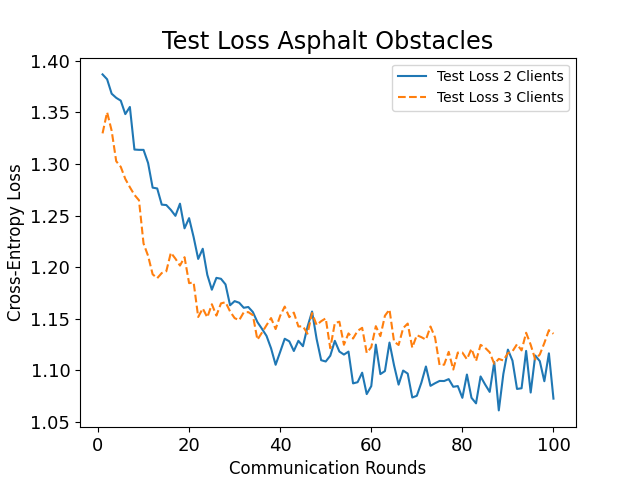}}
    \caption{Performance of asphalt obstacles data set.}
    \label{fig:performance_ashpalt_obstacles}
\end{figure*}

\begin{figure}[h!]
    \centering
    \includegraphics[width = 8.7cm]{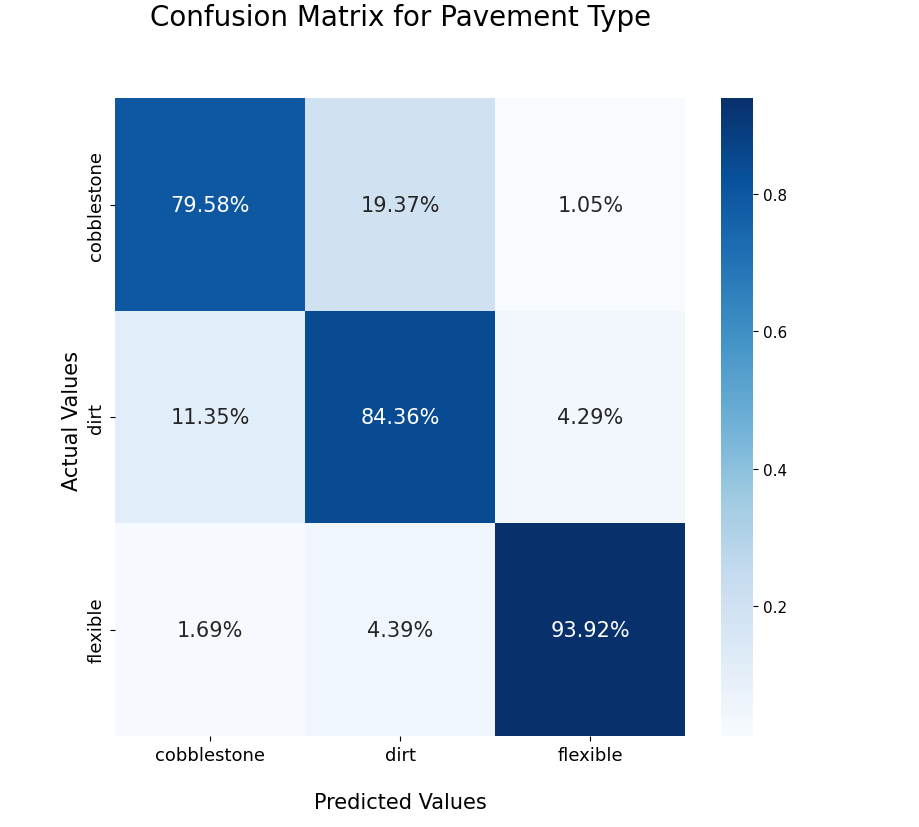}
    \caption{Confusion matrix for pavement type data set. }
    \label{fig:confusion_matrix_pavement_type}
\end{figure}

The confusion matrix in Figure \ref{fig:confusion_matrix_pavement_type} shows that data representing travel over \textit{cobblestones} are hardest for TabNet to predict, incorrectly classifying 20.42$\%$ of all examples. This can be compared to 15.64$\%$ for \textit{dirt roads} and 6.08$\%$ for \textit{flexible roads}. Lastly, TabNet finds distinguishing cobblestone from flexible pavement, and vice versa, the easiest. It only missclassifies these examples as the counterpart in 1.05$\%$ and 1.69$\%$ of the time.  
\subsection{Asphalt Obstacles}\label{subsec:asphalt_obstacles(RESULTS)}
\begin{figure}[h!]
    \centering
    \includegraphics[width = 9.8cm]{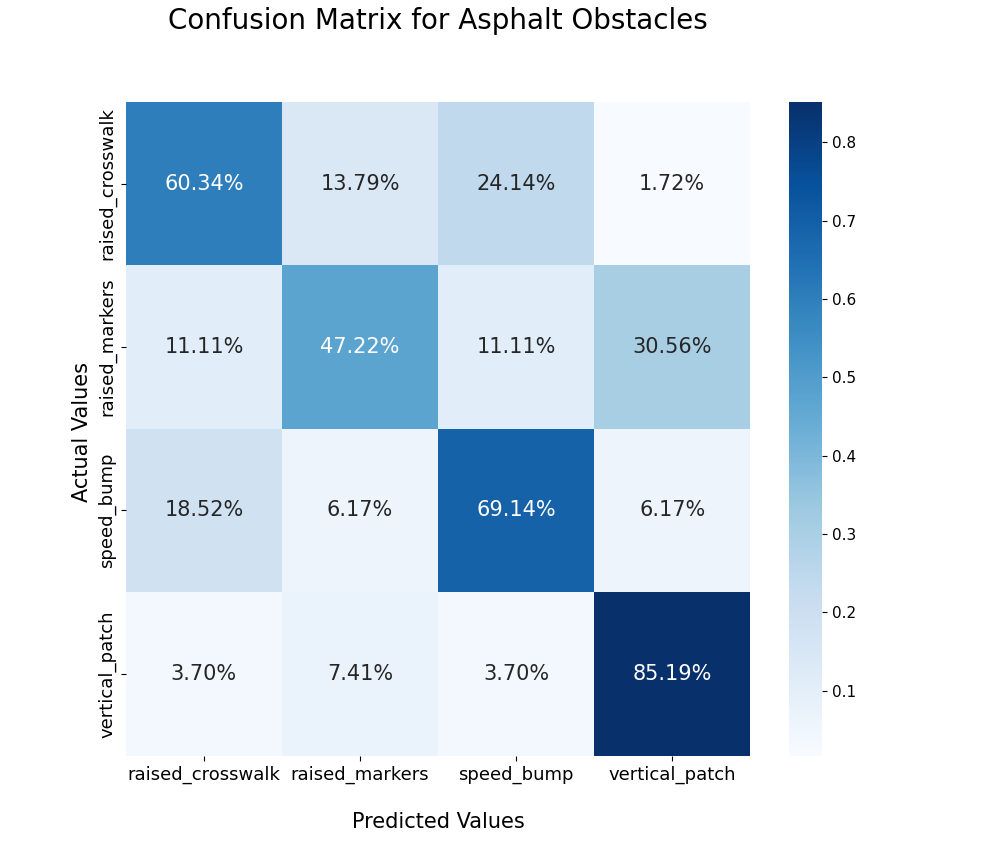}
    \caption{Confusion matrix for asphalt obstacles data set.}
    \label{fig:confusion_matrix_asphalt_obstacles}
\end{figure}

The asphalt obstacles and final data set includes four classes. It is less balanced than the asphalt regularity data set but more balanced than the pavement type data set (see Table I). Again, we plot a confusion matrix, using the predictions from one client in a two-client set-up. We also illustrate the test accuracy and cross-entropy loss of the given number of clients in Figure \ref{fig:performance_ashpalt_obstacles}.
Maximum test accuracy of 68.0$\%$ is reached with two clients. For three clients, maximum test accuracy is 64.5$\%$. The confusion matrix (Figure \ref{fig:confusion_matrix_asphalt_obstacles}) shows that TabNet finds raised markers hardest to predict, followed by raised crosswalk. Moreover, TabNet mistakes raised markers for vertical patches in almost a third of the examples, raised crosswalks for speed bumps in a fourth of the examples, and speed bumps for raised crosswalks in a fifth of the examples. At the same time, TabNet is 85.19$\%$ accurate when predicting vertical patches, only mistaking them for raised crosswalks and speed bumps in 3.70$\%$ of the time, and for raised markers in 7.41$\%$ of the time. 

\section{Discussion}\label{sec:Discussion}
The first observation we make is that TabNet performs well together with FL for all data sets. We clearly identify a convergence of performance as the local training sets grow, and as we communicate back and forth between the server and clients. Furthermore, we present a method using feature extraction and another, combining it with FL and TabNet. The importance of features could also serve as an interesting study. The feature extraction step in our framework is suitable for time series data that shall be converted to tabular data. This step can be skipped if the original data set consists of tabular data. 

For pavement type and asphalt regularity data sets, we find that the results are positively surprising. We argue that the large variation for test accuracy in asphalt regularity data set stems from that the training sets are relatively small in the beginning. After approximately 40 rounds, when the pool set is almost empty, it starts to straighten out. In the pavement type data set, we see that TabNet combined with FL finds it most difficult to predict cobblestones and dirt roads. Intuitively, this is reasonable as both roads can be seen as quite "shaky". This is also in line with the results from \cite{souza2018asphalt}. 

We find that the maximum test accuracy for the asphalt obstacles data set is significantly lower than the results presented in \cite{souza2018asphalt}. This could be due to sub-optimal choice of TabNet hyperparameters. The ablation study conducted by \cite{arik2021tabnet} specifies values for hyperparameters for larger data sets $>$ 10k instances. Our choices could therefore be sub-optimal and can be studied further. Moreover, the data set is quite small (only 781 time series). FL generally requires quite large data sets, and we see that for the first two larger data sets we perform relatively similar to other studies \cite{souza2018asphalt, middlehurst2020canonical}. Additionally, the two classes that were hardest to predict (raised markers and raised crosswalk) were the classes with the least instances in the data set. In this case, there can be a need to handcraft a few distinct features that can support TabNet in its classification and/or conduct a feature analysis study. Between raised markers and vertical patches there seems to be some strong correlation between feature significance. They are both each other's class which is misclassified the most. In more than 30$\%$ of the cases, TabNet mistakes raised markers for vertical patches. Studying Figure \ref{fig:obstacles}, one could argue that raised markers might share most characteristics with vertical patches e.g. height. Nevertheless, we acknowledge that TabNet can predict vertical patches well (85.19$\%$). 

The data sets are interesting to us, as we reason that ICVs should make optimal decisions about what road to take for an arbitrary trip that can maximize user comfort. Many cooperating ICVs can quickly learn characteristics of roads and additional anomalies (obstacles). This information is arguably useful for ICVs as they can choose to avoid uncomfortable roads, and potentially reduce trip duration. Additionally, by applying FL, we can reduce overhead communication and increase privacy by not transmitting trip data that can be used to derive the location of an operating vehicle. 

\section{Related Work}\label{sec:Related_Work}
In the original study, \cite{souza2018asphalt} propose a combination of distance measures between time series to help classify them using a one-nearest-neighbor classifier. They successfully apply it in a one-vehicle setting but do not consider that vehicle can operate together as many ICVs. Further studies have come to show improved performance on the same data sets, with tweaked or novel algorithms, but similarly neglect the integration of FL \cite{middlehurst2020canonical, kannan2021budget}. 

In \cite{wang2021electricity}, they emphasize the importance of principal component analysis (PCA) and feature extraction from time series. Nevertheless, they engineer the features themselves which is a time-consuming task. Our framework proposes extracting a large set of statistical features using the package presented in \cite{christ2018time}. \cite{del2021active} showed how the work from \cite{christ2018time} can be included in time series classification for predictive maintenance in IoT. They also included an active learning sampling strategy which increased the convergence speed. However, they did only consider centralized machine learning applications and not FL. In another work, related to vehicle operations, \cite{thorgeirsson2021probabilistic} showed how a two-scale regression model could be integrated with FL to predict energy demand for an arbitrary route of a vehicle. They even demonstrated how personalization added to the model by clustering groups of drivers. Nevertheless, they do handcraft the features, that the algorithm requires, themselves and their data set is quite small. 

There is much research on how to integrate deep neural networks into DTs. Neural DTs \cite{wang2017using, kontschieder2015deep} use differentiable decision functions, instead of non-differentiable axis-aligned splits. Neural DTs do however not include automatic feature selection which can lead to worse performance. In their study, \cite{sjoberg2019federated} demonstrated how Neural DTs can be combined with FL for the MNIST data set. In collaboration with Volvo, \cite{zhang2021af} extended the work on FL using Neural DTs and showed how it can be applied in a vehicular setting for BDD100k data set. Interestingly, their algorithm outperforms centralized learning. The data are nevertheless images and videos and therefore they disregard time series data.  

Similarly to deep neural DTs, researchers have also developed federated algorithms of random forests \cite{liu2020federated}, conventional DTs \cite{li2020practical}, SVM \cite{hsu2020privacy}, and XGBoost \cite{le2021fedxgboost}. The non-federated implementations for these algorithms have nevertheless been shown to perform worse than TabNet \cite{arik2021tabnet} on large data sets, which is why we seek to develop our own implementation of federated TabNet. To our understanding, there is no work comparing these FL algorithms on standardized data sets, which therefore would be an appropriate study in the future. \cite{zhang2021survey} presents existing work on FL and mentions tree structure implementations for FL. Nevertheless, no superior algorithm is concluded, they only mention the work of \cite{cheng2021secureboost}. 

\section{Conclusion}\label{sec:Conclusion}
TabNet has shown superior performance compared to established models, but combination with FL has not been considered. To our understanding, we are the first to analyze TabNet with FL for tabular data. Moreover, we demonstrate how FL can be applied to the data sets from \cite{souza2018asphalt}. 
Our framework, using the combination of TabNet and FL, is successfully applied to the data sets and shows promising results. 
We include a feature extraction step in our framework to convert the time series data to tabular data. Furthermore, we acknowledge a slight decrease in test accuracy compared to other relevant studies \cite{souza2018asphalt, middlehurst2020canonical}. Our test accuracies vary between $68.0-93.6\%$ and the most difficult data set is the \textit{asphalt obstacles} data set. Nevertheless, in the \textit{asphalt obstacles} data set, we observe that for vertical patches, TabNet classifies the instances 85.19$\%$ correctly. FL does not rely on raw data transmission and supports the increasing demand for privacy preserving operations for edge devices. We also argue for why FL applied to the used data sets is important for ICVs.

In the future, we shall tune the hyperparameters for the smaller data sets at hand and evaluate different FL optimization algorithms performance on similar data sets. Furthermore, we shall investigate the combination of TabNet and FL on other benchmark data sets and extend our work to include regression tasks and not only use it for classification. In such a work, we aim to compare federated TabNet with current state-of-the-art federated algorithms that can be utilized for tabular data. Lastly, we shall compare TabNet and FL with a centralized approach and study the difference in terms of overall performance and communication efforts.  

\section*{Acknowledgement}\label{sec:acknowledgment}
This work was partly supported by the TRANSACT project. TRANSACT (https://transact-ecsel.eu/) has received funding from the Electronic Component Systems for European Leadership Joint Under-taking under grant agreement no.  101007260.  This joint undertaking receives support from the European Union’s Horizon 2020 research and in-novation programme and Austria,  Belgium,  Denmark,  Finland,  Germany,Poland, Netherlands, Norway, and Spain.

\bibliographystyle{IEEEtran}
\bibliography{main.bib}

\end{document}